\documentclass[10pt]{article} 
\usepackage[preprint]{tmlr}

\usepackage{amsmath,amsfonts,bm}

\def\eqref#1{equation~\ref{#1}}

\def\1{\bm{1}}

\DeclareMathAlphabet{\mathsfit}{\encodingdefault}{\sfdefault}{m}{sl}
\SetMathAlphabet{\mathsfit}{bold}{\encodingdefault}{\sfdefault}{bx}{n}

\usepackage{hyperref}
\usepackage{url}
\usepackage{graphicx}
\usepackage{booktabs}
\usepackage{amsmath}
\usepackage{amssymb}
\usepackage{multirow}

\title{Input Pathways Shape Few-Shot, Not Zero-Shot, Binding in Tiny Transformers:
A Fully-Enumerable Study}

\author{\name Yoshiyuki Ootani \email info@ootanl.com \\
      \addr Independent Researcher}

\newcommand{\microground}{\textsc{MicroGround}}

\def\month{07}
\def\year{2026}
\def\openreview{\url{https://openreview.net/forum?id=XXXX}}

\begin{document}
\maketitle

\begin{abstract}
How does the \emph{way} information reaches a transformer---as symbolic in-context tokens, as a
clean per-factor ``oracle'' code, or as an entangled perceptual vector---affect whether the model
can \emph{bind} that information compositionally? We study this in a deliberately extreme regime:
transformers of $\sim$6--10K parameters on finite factored worlds that we enumerate exhaustively,
so every behavioral measurement is evaluated on the entire input space (zero sampling variance),
the informative input channels are matched in the sense that each uniquely determines the answer
(verified by exact Bayes ceilings), and every headline comparison is run at 10--20 seeds with
paired nonparametric tests (robustness arms use 5--20).
We report four findings. (1)~\emph{Endpoint invariance:} on held-out binding queries no informative
route yields converged zero-shot composition---each ends at or below chance despite an exact Bayes
ceiling of $1.0$, so within our bounded sweep (a ${\sim}25\times$ parameter increase, a $10\times$
learning-rate range, weight decay, and longer training) all routes converge to lookup-like solutions
despite information-sufficient inputs; this also holds under a second injection scheme, in which
the state occupies its own sequence position rather than being added to every token
(\S\ref{sec:prefix}). (The ceiling certifies that the input determines the answer,
not that the held-out mapping is identifiable from a training objective for which lookup suffices.)
(2)~\emph{Readability governs few-shot binding:} when a small fraction of the held-out query type
is leaked into training, sample efficiency within the tested cells is predicted by the
\emph{readability} of the code. A poorly readable entangled code, whose raw linear decodability
drops $0.95 \to 0.58$, is the worst route and saturates far below the readable alternatives
($-0.147$ pooled against the most readable route, under both injection schemes we tested); a
dimension-matched control and a graded readability sweep---few-shot accuracy is monotone in
decodability at fixed input dimension---isolate readability from input dimension; and the effect
replicates on a second held-out query type at 20 seeds and in a modestly larger three-object world.
(3)~\emph{The injection site, not the input pathway:} v1 read the remaining ordering among readable
routes---shared projections above the clean per-factor oracle---as input-pathway parameter sharing.
It is neither sharing nor a pathway property. The compared pathways are one function class, so
sharing was never manipulated; the ordering is driven by the initialization scale of the state that
broadcast-addition injects, and it vanishes when the state is given its own sequence position
($+0.057 \to -0.006$, $n{=}20$ seeds/cell paired by initialization seed). We report this as a
cautionary result about an unexamined implementation default rather than a finding about pathways. (4)~\emph{Failure anatomy:} the symbolic route fails by
\emph{losing} the answer at the readout position; index routes fail by \emph{systematically
mis-binding}---the answer stays decodable in the residual, yet a direct input intervention shows
the converged output is more sensitive to the \emph{wrong} object slot than to the correct one---and
the entangled route inherits, rather than improves on, the readability of its input.
Of these, the central positive claim is readability; endpoint invariance and failure anatomy are
diagnostic constraints on it, and the injection-site result marks the boundary of what an
input-pathway comparison can establish.
All code, manifests, and per-seed logs are available for exact reproduction.
\end{abstract}

\paragraph{Note on this version (v2).} This version corrects the first of the two
few-shot factors claimed in v1. v1 attributed the few-shot advantage of shared input pathways to
\emph{input-pathway parameter sharing}; that attribution is withdrawn, for two independent reasons
established after v1. (i)~On disjoint one-hot blocks a shared linear projection is
block-decomposable, so the shared, untied, and per-factor pathways compared here lie in a single
function class (trainable pathway parameters $408/480/384$; the shared pathway is the per-factor one
plus a $24$-dimensional bias). The oracle-vs.-one-hot-shared contrast therefore never manipulated
sharing. (ii)~Direct manipulation locates the surviving difference in the \emph{initialization
scale} of the injected state, and a matched sweep that gives the state its own sequence position,
instead of adding it to every token, removes it ($+0.057 \to -0.006$;
\S\ref{sec:prefix}). The effect is a property of broadcast-additive conditioning, not of input
pathways. We also withdraw the above-chance zero-shot transient and the double dissociation built on
it: the ``Peak'' column of Table~\ref{tab:zeroshot} is a mean of per-seed maxima, a selection
statistic (\S\ref{sec:dynamics}). What stands is readability, which the injection site leaves
unchanged ($-0.147$ under both schemes; \S\ref{sec:fewshot}), and endpoint invariance, which prefix
routing strengthens. Superseded claims are marked as such in place rather than deleted, so that this
version can be read against v1; \S\ref{sec:prefix} reports the new sweep in full.

\section{Introduction}
\label{sec:intro}
A transformer can receive the same fact in different ways. ``The object on the right is a square''
can arrive as words in the token stream, as a slot in a structured state vector, or as coordinates
of an entangled ``perceptual'' embedding. The symbol grounding debate
\citep{harnad1990symbol,bender2020climbing,bisk2020experience,merrill2021provable} asks whether the
non-symbolic routes confer something the symbolic route cannot; recent mechanistic work localizes
grounding-like mechanisms in language models \citep{wu2025mechanistic}, and a related line shows
that \emph{where} matched information is stored (weights versus context) changes how a transformer
generalizes from it \citep{chan2022transformers}. But these studies never hold the information
content of the routes constant: symbolic and perceptual conditions differ in dataset, model, and
difficulty, so route is confounded with everything else.

We ask the controlled version of the question. Fix a finite world of independent categorical
factors; deliver the \emph{same} state---provably sufficient to answer every query---through five
different input routes; and measure, exhaustively, what the model learns, how it learns it, and how
it fails. Our testbed, \microground{}, is small enough ($128$--$256$ states, $\sim$6--10K-parameter
transformers) that we can evaluate every model on the \emph{entire} query space, compute exact
Bayes-optimal ceilings per route, and probe every checkpoint. This trades scale for a property
larger studies cannot have: when a model fails, we can say \emph{exactly} whether the information
was absent, present-but-unread, or present-but-unusable, and when two routes differ, no sampling
noise or information mismatch can explain the difference. We therefore study \emph{information-matched
input encodings and pathways}, not grounding in the rich sense: the ``perceptual'' routes are fixed
synthetic codes of enumerated categorical states, and grounding is a motivating question, not a
claim we test.

Our target phenomenon is \emph{compositional binding}: answering ``shape of right'' requires
binding the queried attribute to the correct object slot \citep{greff2020binding,feng2024how}. We
hold out an entire query type (e.g., the model is never asked the shape of the right object during
training) and test whether each route composes the attribute concept, which it has seen, with the
object slot, which it has also seen. We complement binding with a systematic transformation task
(counterfactual successor, $v \mapsto (v{+}1) \bmod k$) under a transition holdout, an instance of
the leave-region-out diagnostic of \citet{hu2024case}.

An initially plausible hypothesis---that entangled, grounded-like routes encourage compositional
structure---is falsified here by a strong-entangled control (an entangled perceptual code that
defeats linear decoding), replicated across two held-out shape query types. What the controls isolate
instead is a mechanistic account of \emph{which properties of an input pathway matter, for what, and
why}, together with an unusually well-controlled negative result about zero-shot
compositionality at this scale.

\paragraph{Contributions.}
\begin{itemize}
\item \textbf{Testbed and method} (\S\ref{sec:testbed}--\S\ref{sec:ceilings}): \microground{},
fully-enumerable factored worlds with route-factored conditions (symbolic / factored-oracle /
weak- and strong-entangled perceptual / one-hot-shared, plus floor and scrambled controls), exact
per-route solvability ceilings, exhaustive evaluation with zero sampling variance, and a
statistical protocol (converged-accuracy metric, 5--20 seeds with 10--20 for all headline
comparisons, paired Wilcoxon with Holm--Bonferroni, bootstrap CIs, effect sizes) under an explicit
exploratory/confirmatory split.
\item \textbf{Endpoint invariance} (\S\ref{sec:endpoint}): no informative route achieves converged
zero-shot compositional binding; robust to capacity, weight decay, and training length; all failures
occur strictly below a ceiling of $1.0$, so the inputs are information-sufficient while the held-out
mapping is not identifiable from a training objective for which lookup suffices. (A transition-holdout
diagnostic shows the same memorization pattern; we treat it as weaker supporting evidence, as the
tiny world may not reward the rule.)
\item \textbf{Readability predicts few-shot binding} (\S\ref{sec:fewshot}): practical readability
of the input code is the one property of the input that predicts sample efficiency here
($-0.147$ pooled against the most readable route, in both injection schemes), isolated from input
dimension by a dimension-matched control and by a graded sweep in which few-shot accuracy is
monotone in decodability; replicated across two held-out shape query types at 20 seeds, corroborated
as a stress check in a modestly larger three-object world, and robust to description order and to
the synthetic mix.
\item \textbf{The injection site is a confound in route comparisons} (\S\ref{sec:prefix}): the
ordering among readable routes that v1 attributed to pathway sharing survives neither of two checks.
The pathways compared are one function class (so sharing was never manipulated), and the residual
effect---the initialization scale of the injected state---disappears when the state is given its own
sequence position. A route comparison therefore has to control where conditioning information
enters, not only what it contains.
\item \textbf{Failure anatomy} (\S\ref{sec:anatomy}): control-task probes plus a causal
input-intervention separate three failure modes---representation death at the readout position
(symbolic), systematic mis-binding with the answer decodable but the output depending, under input
intervention, on the wrong slot (index routes), and no readability gained over the input (entangled route).
\end{itemize}

\section{Related Work}
\label{sec:related}
\paragraph{Mechanistic symbol grounding.} Closest to our framing, \citet{wu2025mechanistic} train
transformers and SSMs from scratch on a testbed where each lexical item has an environmental token
and a linguistic token, quantify grounding as the surprisal reduction of the linguistic token given
a \emph{matched} versus \emph{mismatched} referent, and localize a middle-layer ``aggregate''
attention mechanism. Their manipulation is referent correctness (matched vs.\ mismatched referent
within a from-scratch token testbed, with a multimodal-dialogue replication), not \emph{route}, and
the conditions are never information-matched, so a similarity between conditions cannot separate
route from modality difficulty. Holding information constant across routes is exactly our
manipulation---and our falsified hypothesis (\S\ref{sec:fewshot}) shows why the control matters.

\paragraph{Where information is stored, and how it is encoded.} \citet{chan2022transformers} show,
via a partial-exposure paradigm, that matched information generalizes rule-like from weights but
exemplar-like from context. We fix the storage locus (all state routes are in-context side-channels)
and vary the \emph{encoding} instead; but the connection is more than orthogonality: their
context$\to$exemplar bias arguably \emph{predicts} our endpoint invariance (all in-context routes
collapse to memorization, \S\ref{sec:endpoint}), which we take as convergent support for the
inductive-bias reading rather than an independent finding. That the \emph{form} of an encoding
governs generalization is also known on other axes---most directly positional-encoding choice and
length generalization \citep{kazemnejad2023impact}---and compositional generalization has a
standing taxonomy \citep{hupkes2020compositionality}; our contribution is an information-matched
route manipulation with exact ceilings, not the bare claim that encoding matters.

\paragraph{Rule versus memorization.} \citet{hu2024case} introduce leave-region-out diagnostics on
arithmetic grids and show GPT-2-scale transformers reason case-based rather than rule-based; our
transition holdout is a direct instance of their diagnostic, which we credit as the method. Our
contribution there is the route comparison and the exact-ceiling framing (models fall below even
the blind-majority bound). Grokking work \citep{power2022grokking,nanda2023progress,zhong2023clock}
motivates our weight-decay arm, which fails to induce rule learning here (\S\ref{sec:endpoint}).

\paragraph{Binding.} \citet{feng2024how} characterize the additive Binding-ID mechanism by which
large language models bind entities to attributes in context; their setting is purely symbolic.
\citet{greff2020binding} frame binding as a core representational problem. We study the same
computational problem at a scale where the full input space is enumerable and ask how the binding
substrate depends on the input pathway.

\paragraph{Exact bounds and toy-model interpretability.} \citet{gross2024compact} prove
computer-assisted \emph{lower} bounds on a specific trained model's accuracy, priced by the
compactness of its circuit; Tracr \citep{lindner2023tracr} compiles known circuits. Our solvability
analysis is a different object: model-agnostic \emph{upper} bounds (Bayes ceilings) computed a
priori from the enumeration, used to classify failures as information- versus inductive-bias-driven.

\paragraph{World models, compositional generalization, and encodings.} Emergent world
representations are probed and edited in task-defined worlds \citep{li2023emergent}; grounded
compositional benchmarks \citep{lake2018generalization,ruis2020benchmark,kim2020cogs} are large and
behavior-only. Two prior results are especially close and temper our positive claims. First,
architectural weight/parameter \emph{sharing} is known to aid systematic generalization
\citep{csordas2021devil}; v1 read our oracle-vs.-one-hot-shared contrast as an
information-matched instance of this, and that reading is withdrawn in this version---on disjoint
one-hot blocks the two pathways are one function class, so the contrast varies no sharing
parameter (\S\ref{sec:fewshot}). Second, \citet{montero2021role} show
that disentangled representations, even when handed to the model, are \emph{not sufficient} for
harder compositional generalization---the direct precedent for our ``clean oracle is worst''
result. v1 offered pathway sharing as the mechanism; this version withdraws that and locates the
ordering in the injection site instead (\S\ref{sec:prefix}), so what we add here is a
demonstration that the ordering is not a property of the code at all. Our finding that models
exploit linearly-readable features and fail to disentangle further connects to shortcut learning
and simplicity bias \citep{geirhos2020shortcut,shah2020pitfalls,hermann2020shapes} (with the
caveat that readability here \emph{helps} rather than names a failure mode) and to the difficulty
of unsupervised disentanglement \citep{locatello2019challenging}; our probes follow the control-task methodology of
\citet{hewitt2019designing}, with representation geometry in mind \citep{elhage2022toy}. Grounding
debate anchors \citep{bender2020climbing,bisk2020experience,merrill2021provable,patel2022mapping}
motivate the operational question without settling it.

\paragraph{What is isolated here, versus already known.} Individually, several ingredients are not
new: parameter sharing aids systematic generalization \citep{csordas2021devil}, disentangled codes
are not sufficient for it \citep{montero2021role}, and small transformers memorize under weak
compositional pressure. Our contribution is not any single effect but the \emph{confound removal}
that lets these be separated: by matching information exactly (verified by per-route Bayes
ceilings), enumerating the entire input space, and factoring the input pathway into code format,
readability, and---added in this version, after it proved to be the variable that mattered---the
site at which the state is injected, we isolate variables that modality comparisons ordinarily
leave entangled with dataset, capacity, difficulty, and information content.

\section{The \microground{} Testbed}
\label{sec:testbed}
\paragraph{Worlds.} The single-object world has four independent categorical factors (colour,
shape, position, size) of cardinalities $(4,4,4,2)$: $128$ states. The two-object world places two
objects in fixed positional roles (left, right), each with a colour and a shape: $4^4 = 256$
states. Both worlds are enumerated in full; every evaluation below is over the \emph{entire}
relevant query space, so for a fixed split and trained model, evaluation has zero sampling variance;
across the reported runs, variance comes from model initialization and, where applicable, the split
or leakage seed.

\paragraph{Tasks.} \emph{Attribute} (report a queried factor; a lookup used as a sanity ceiling),
\emph{counterfactual} (report the successor $(v{+}1) \bmod k$ of a queried factor), and
\emph{binding} (report one attribute of the object at a queried position, e.g.\ ``shape of
right''). The two-object binding task has four query types, one per (position, attribute) pair,
$256$ queries each.

\paragraph{Routes (conditions).} A condition factors into a text form and a state encoder, so the
same exhaustive query set is realized under every route. The five informative routes each uniquely
determine the answer (verified in \S\ref{sec:ceilings}); the two floors
(\texttt{text\_minimal}, \texttt{uninformative\_state}) provide none of the state, and
\texttt{scrambled\_state} is an information-matched control code. The same exhaustive query set is
realized under every route (Table~\ref{tab:routes}).
\begin{table}[h]
\centering\footnotesize
\resizebox{\textwidth}{!}{%
\begin{tabular}{llll}
\toprule
Condition & Input & State channel & Raw linear dec.\\
\midrule
\texttt{text\_only} (symbolic) & description + question & --- & (in tokens)\\
\texttt{state\_factored} (oracle) & question & per-factor index $\to$ per-factor emb.\ tables & indices\\
\texttt{state\_onehot\_shared} & question & concat.\ one-hot $\to$ shared linear & $1.0$\\
\texttt{state\_perceptual} (weak) & question & fixed 2-layer tanh mix, $d{=}16$ $\to$ shared linear & $0.95$\\
\texttt{state\_perceptual\_hard} & question & fixed 3-layer tanh mix, $d{=}8$, gain $2.5$ $\to$ shared linear & $0.58$\\
\midrule
\texttt{text\_minimal} (floor) & question & --- & ---\\
\texttt{uninformative\_state} (control) & question & constant vector & ---\\
\texttt{scrambled\_state} (control) & question & fixed bijective relabeling & indices\\
\bottomrule
\end{tabular}}
\caption{Routes. ``Raw linear dec.'' is 5-fold logistic decoding of a factor from the raw state
input (mean over factors). The one-hot-shared and weak-perceptual routes use an identical
$16{\to}24$ linear input pathway (exactly matched parameter count); the factored oracle uses
per-factor embedding tables of nearly identical total size. The perceptual mixes are fixed
(run-independent), so they are a shared ``perception,'' not trainable preprocessing. We use
``perceptual'' descriptively: \texttt{state\_perceptual} is a synthetic entangling map standing in
for a distributed non-symbolic code, not a learned or naturalistic perception. Our claims are about
input encodings and pathways; grounding is motivation, not a claim.}
\label{tab:routes}
\end{table}

\paragraph{Generalization splits.} Beyond random $80/20$ splits we use: \emph{binding holdout}
(remove an entire (position, attribute) query type from training; test on all $256$ of its
queries), \emph{$k$-shot binding} (leak a fraction $f$ of the held-out type back into training;
test on the rest), and \emph{transition holdout} (remove a random $30\%$ of (factor, source-value)
successor transitions; after \citealp{hu2024case}). Splits partition states or query types; query
generation is otherwise exhaustive and deterministic. The model is a 1-layer (2-layer in the
capacity arm) pre-norm transformer, hidden size $24$ ($\sim$6--10K parameters), trained with AdamW;
the state embedding is added to every token position.

\section{Protocol: Exhaustive, Converged, and Replicated}
\label{sec:protocol}
Three methodological choices matter for everything below. \textbf{(i) Converged, not peak.} We
report the \emph{converged} test accuracy (mean over the last $10\%$ of evaluations). During
exploration we found that the peak over training inflates transient generalization spikes that
decay by convergence (\S\ref{sec:dynamics} turns this artifact into an object of study); flat
controls (collapse $= 0.000$) rule out max-of-noise inflation. \textbf{(ii) Balanced accuracy with
baselines.} The primary metric is balanced accuracy (mean over per-query-type accuracy), always
reported against the chance and majority baselines; because evaluation is exhaustive, both are
exact. \textbf{(iii) Replication discipline.} Initialization and split seeds are separated; headline
results use $10$--$20$ seeds (robustness arms $5$--$20$) with bootstrap $95\%$ CIs, paired
Wilcoxon signed-rank tests across seeds, Holm--Bonferroni correction, and Cliff's $\delta$. Effects discovered during exploration (on
held-out type ``shape of right,'' \texttt{bind:3}) were re-tested at 20 seeds and then replicated
on a fresh held-out type (``shape of left,'' \texttt{bind:1}) with all routes run once,
confirmatory-style; we mark exploratory versus confirmatory results throughout. Every run appends
a self-describing record (config, seeds, trajectories) to a JSONL manifest from which all tables
and figures re-aggregate without retraining.

Two scope notes on the statistics. \emph{Interpreting null endpoint differences.} The endpoint-invariance
and no-transition-difference claims are acceptances of the null; we read them as bounded (``no knob
in this box''), and note that the paired zero-shot route-difference confidence intervals are smaller
than the $+0.05$--$0.08$ few-shot effects in Table~\ref{tab:fewshot}; we therefore treat endpoint
invariance as a bounded empirical observation, not a formal
equivalence claim.
\emph{The mechanism arms use $n{=}10$}; their effect sizes are large ($|\delta|\ge0.6$, $p<10^{-3}$)
and the one place a small effect mattered---the k-shot ordering---is exactly where our $10\to20$-seed
firm-up caught and shrank it, so we trust the large mechanism effects at $n{=}10$ while reporting the
smaller ones (symbolic route) as non-significant.

Two exploratory refinements did not survive this protocol. A three-way
route ordering visible at 10 seeds shrank to a two-group structure at 20 seeds; and our headline
hypothesis---that a grounded side-channel induces compositional structure---was falsified by the
strong-entangled control we designed to test it (\S\ref{sec:fewshot}).

\section{Exact Solvability Ceilings}
\label{sec:ceilings}
Because the worlds are finite we compute, for every (task, split, route), the Bayes-optimal
balanced accuracy on the test queries: group test queries by exact input identity---two states
yield the same group only if their input vectors are identical, an injectivity check over the
finite enumeration---and predict the within-group majority. This is an a priori, model-agnostic
\emph{upper} bound on any learner's test behavior \emph{given the input}---a channel property, to
be contrasted with proof-based \emph{lower} bounds for specific trained networks
\citep{gross2024compact}. Two things it does \emph{not} certify: that the mapping is identifiable
from \emph{training} (for a held-out query type it is not), and that the input is easy for a
particular decoder or model class to read.

The audit yields three facts used throughout. (1)~Every informative route is \emph{injective}
(ceiling $1.0$) on every split: distinct states yield distinct inputs, so no informative route below fails for
lack of information in the input---failures are of inductive bias or (for held-out query types)
train-time identifiability, not of channel capacity. (2)~The floors are exactly chance on binding
holdouts (ceiling $0.250$: the question alone carries nothing) and $0.833$ on transition holdouts
(held-out transitions have degenerate target distributions a blind majority guesser can exploit).
(3)~The controls behave as designed: \texttt{uninformative\_state} tracks \texttt{text\_minimal}
everywhere (e.g., $0.296$ vs.\ $0.296$ peak on the random-split counterfactual task; Cliff's
$\delta = +0.02$), and \texttt{scrambled\_state} reaches $1.000$ in distribution---an arbitrary
bijective code is learned perfectly when information content is matched. Full ceiling values are
in Appendix~\ref{app:ceilings}.

\section{Results I: Endpoint Invariance---No Route Yields Zero-Shot Compositionality}
\label{sec:endpoint}
\paragraph{Binding.} Table~\ref{tab:zeroshot} gives converged accuracy on the held-out query type
(exploratory holdout, $n{=}20$). Every route ends at or below chance ($0.25$); the strong-entangled
route ends far below ($0.146$), and the oracle ends below ($0.198$)---systematic error, not
agnosticism (\S\ref{sec:anatomy}). In-distribution learning is intact (all informative routes fit
the three trained query types; random-split sanity checks confirm the models learn the task in
distribution), so the failure is specific to composition. The endpoint is also invariant to the
injection site: in a matched sweep that gives the state its own sequence position instead of adding
it to every token embedding, all four informative state routes again end at or below chance
($0.189$ oracle, $0.192$ one-hot-shared, $0.224$ weak-perceptual, $0.167$ strong-entangled;
$n{=}20$ each, \S\ref{sec:prefix}). \textbf{Scope.} Because training
covers three of the four query types, pure lookup is a low-loss training solution, so the world
does not strongly \emph{reward} composition; ``endpoint invariance'' should be read as ``in
training regimes where memorization suffices, no informative route within our bounded sweep (${\le}2$ layers,
${\le}96$ hidden, AdamW, LR/wd ranges below, ${\le}4000$ epochs) escapes it,'' not as a universal
limit. The exact ceiling of $1.0$ rules out an \emph{information} explanation, not a
training-objective one; a design that makes lookup costly (more held-out types) is the right next
test and is future work.

\begin{table}[h]
\centering\small
\begin{tabular}{lccc}
\toprule
Route & Converged held-out acc.\ [95\% CI] & Peak of mean (epoch) & Collapse\\
\midrule
\texttt{text\_minimal} (floor) & 0.237 [0.21, 0.25] & 0.237 & 0.000\\
\texttt{uninformative\_state} & 0.237 [0.21, 0.25] & 0.237 & 0.000\\
\texttt{text\_only} & 0.227 [0.19, 0.26] & 0.272 (10) & 0.131 [0.09, 0.18]\\
\texttt{state\_factored} & \textbf{0.198} [0.16, 0.23] & 0.211 (20) & 0.026\\
\texttt{state\_onehot\_shared} & 0.215 [0.20, 0.23] & 0.240 (10) & 0.034\\
\texttt{state\_perceptual} & 0.224 [0.20, 0.24] & 0.265 (5) & 0.084 [0.06, 0.12]\\
\texttt{state\_perceptual\_hard} & \textbf{0.146} [0.11, 0.18] & 0.255 (10) & 0.140 [0.10, 0.19]\\
\bottomrule
\end{tabular}
\caption{Zero-shot compositional binding (held-out query type ``shape of right''), $n{=}20$ seeds,
chance $= 0.250$, ceiling of every informative route $= 1.0$. No route generalizes at convergence,
and two end significantly \emph{below} chance (bold; $p_{\text{adj}} = 2.5\times10^{-4}$ for the
oracle and $0.030$ for one-hot-shared), which \S\ref{sec:anatomy} traces to systematic mis-binding.
\textbf{Correction in v2:} v1's third column was the mean of per-seed \emph{maxima} ($0.358$,
$0.308$ and $0.286$ for the symbolic, weak- and strong-perceptual routes), a statistic that selects
each seed's best epoch and is biased upward under the null. The column now reports the peak of the
across-seed mean, with every seed read at the same epoch; the selection inflation it removes is
$+0.086$ for the symbolic route and $+0.043$\,/\,$+0.031$ for the perceptual ones. At these epochs
no route is above chance (symbolic $p = 0.25$, weak-perceptual $0.58$, strong-entangled $0.83$), so
the above-chance transient claimed in v1 is withdrawn (\S\ref{sec:dynamics}).}
\label{tab:zeroshot}
\end{table}

\paragraph{Transitions.} On the counterfactual successor task with $30\%$ of transition types held
out, every route converges to $0.000$--$0.058$ on held-out transitions (weak-perceptual exactly
$0.000$; no significant route differences, Holm $p \ge 0.72$; $n{=}10$; the training set is fully
memorized, all-space balanced accuracy $0.69$--$0.75$, flat for the last $960$ of $1000$ epochs).
Note that even the best of these, $0.058$, falls far below $0.833$: every route is worse than a
blind majority guesser---the models do not abstain but emit systematically wrong answers. We
caution, however, that with only three visible transitions per factor the world may not
\emph{reward} the successor rule at all; we read the transition result as memorization under weak
compositional pressure, not as proof of a deep inductive-bias pathology, and do not lean on it
beyond the binding conclusion.

\paragraph{Robustness.} The invariance is not a budget artifact. \emph{Capacity:} at hidden sizes
$24\to48\to96$ and $1\to2$ layers ($6$K$\to$$21$K$\to$$40$K$\to$$153$K parameters, ${\sim}25\times$),
converged held-out binding stays at
or below chance for every route swept (symbolic, oracle, weak-perceptual); capacity instead
softens the failure's distinctive character (the oracle's below-chance converged accuracy rises
from $0.178$ to $0.240$, approaching chance; the symbolic per-seed-max peak decays $0.385 \to 0.258$
in the $n{=}8$ capacity arm, against the $n{=}20$ per-seed-max value of $0.358$ reported in the
caption of Table~\ref{tab:zeroshot}; both are selection statistics and we read the decay as
descriptive only, \S\ref{sec:dynamics}).
\emph{Learning rate:} at $3\times10^{-4}$ and $3\times10^{-3}$ ($10\times$ around the default),
converged held-out binding remains $0.18$--$0.24$ for all three routes tested ($n{=}10$).
\emph{Regularization:} on the oracle route, weight decay $\in \{0.01, 0.1, 0.3, 1.0\}$ for
$4{,}000$ epochs ($\approx$52K steps) leaves held-out transition accuracy at $0.000$ in every
cell---the regularizer that reliably induces grokking in modular arithmetic at this parameter
scale does nothing here, consistent with the successor task's lack of compressive coupling across
values, and with a world (three visible transitions per factor) that may simply be too small to
reward the rule. \emph{Duration:} trajectories are flat long before evaluation ends. Grid details
are in Appendix~\ref{app:robust}.

\section{Results II: The Transient Phase, Reassessed}
\label{sec:dynamics}
Because evaluation is exhaustive, every point of a test trajectory is an exact property of the
network at that epoch, so a trajectory here is a measured object and not a noisy estimate. v1 used
that to claim a transient phase of genuine zero-shot transfer in the distributed codes---the
symbolic route reaching $0.358$ around epoch $12$ and the perceptual routes $0.286$--$0.308$---and
paired it with the few-shot result as a double dissociation. \textbf{We withdraw that claim.} The
column it rested on is a mean of per-seed maxima: each seed contributes its own best epoch, so the
statistic selects on the very quantity being tested and is biased upward even under the null. Read
selection-free, as the peak of the across-seed mean with every seed evaluated at the same epoch, no
route clears chance: the symbolic route peaks at $0.272$ at epoch $10$ (SD $0.082$ across seeds,
$p = 0.25$ against chance), weak-perceptual at $0.265$ ($p = 0.58$), strong-entangled at $0.255$
($p = 0.83$). The selection inflation is $+0.086$ for the symbolic route and $+0.043$\,/\,$+0.031$
for the perceptual ones (Table~\ref{tab:zeroshot}).

Two things do survive. First, the shape of the trajectories differs by code format: the distributed
codes (\texttt{text\_only}, weak and strong perceptual mixes) rise and then fall, while the
index-like codes (\texttt{state\_factored}, \texttt{state\_onehot\_shared}) descend to their
endpoints monotonically---a difference in how routes approach lookup, which is what the collapse
column describes. Second, that approach ends asymmetrically: the two index-like routes converge
significantly \emph{below} chance ($0.198$ and $0.215$, $p_{\text{adj}} = 2.5\times10^{-4}$ and
$0.030$), systematic mis-binding rather than agnosticism (\S\ref{sec:anatomy}), while the
distributed codes end at chance. What does \emph{not} survive is the sharp statistical separation v1
reported between collapse magnitudes, because that quantity is also computed as maximum minus
converged and inherits the same selection: evaluated instead at a fixed epoch, the two pairs v1
separated ($p_{\text{adj}} = 7.8\times10^{-4}$ and $9.0\times10^{-4}$) are no longer distinguishable
($p_{\text{adj}} = 0.19$ for both; Cliff's $\delta = +0.35$ and $-0.49$). We therefore present the
dynamics as a description of how each route reaches its endpoint, not as evidence of a compositional
phase, and the capacity observation as descriptive likewise (the $n{=}8$ capacity-arm per-seed-max
peak decays $0.385 \to 0.258$ from 6K to 153K parameters).

\begin{figure}[t]
\centering
\includegraphics[width=0.98\textwidth]{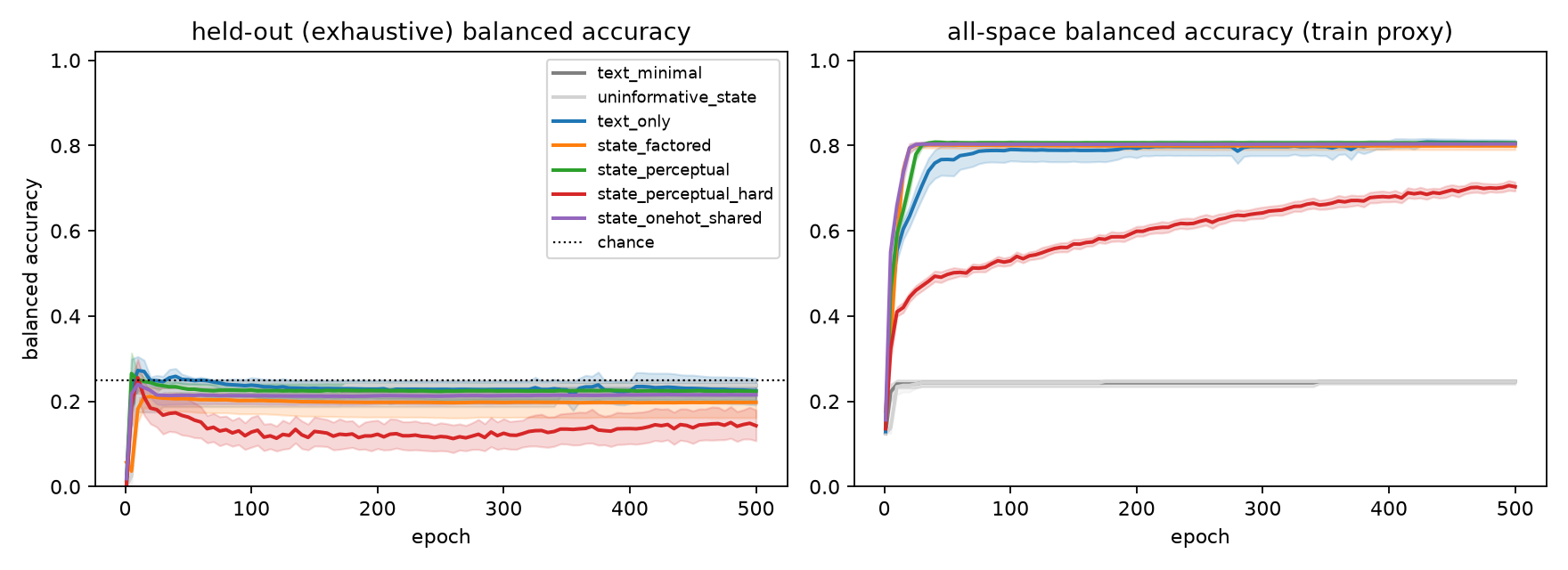}
\caption{Training dynamics by route on the held-out binding type ($n{=}20$ seeds; bands are
bootstrap 95\% CIs; every point is an exhaustive evaluation). Left: held-out (never-trained) query
type. The distributed codes (blue/green/red) rise and fall while the index-like codes
(orange/purple) descend monotonically to below chance; no route's across-seed mean clears chance at
any epoch, and v1's above-chance transient was an artifact of averaging per-seed maxima
(\S\ref{sec:dynamics}). Right: all-space accuracy (train proxy) shows all informative routes
memorize the trained types.}
\label{fig:dynamics}
\end{figure}

\section{Results III: Failure Anatomy---Where the Answer Goes}
\label{sec:anatomy}
For each route we retrained models with mid-training checkpoints ($n{=}10$) and measured, on the
held-out type: behavioral accuracy; the \emph{mis-binding rate} (among states whose left and right
shapes differ, the probability that the model answers with the \emph{left} shape); and a 5-fold
linear probe decoding the correct (right) shape from the final-token residual, with a
control-task probe \citep{hewitt2019designing} subtracted (selectivity $=$ probe $-$ control);
see Table~\ref{tab:anatomy}.

\begin{table}[h]
\centering\small
\begin{tabular}{lcccc}
\toprule
Route (converged) & Behav. & Mis-binding & Probe & Selectivity\\
\midrule
\texttt{text\_only} & 0.241 & 0.510 & 0.278 & $-0.013$ (0.215 at ep5)\\
\texttt{state\_factored} & 0.178 & 0.706 & \textbf{0.952} & 0.687\\
\texttt{state\_onehot\_shared} & 0.217 & \textbf{0.866} & \textbf{0.953} & 0.685\\
\texttt{state\_perceptual} & 0.173 & 0.431 & \textbf{0.963} & 0.682\\
\texttt{state\_perceptual\_hard} & 0.141 & 0.314 & 0.589 & 0.316\\
\bottomrule
\end{tabular}
\caption{Failure anatomy on the held-out binding type ($n{=}10$; chance behav.\ $=0.25$; probe
chance $=0.25$). Three modes: the symbolic route \emph{loses} the answer at the readout position
(selectivity $0.215$ at epoch 5, coinciding with its behavioral transient, decaying to $\approx 0$);
index routes keep the answer perfectly decodable yet systematically read the wrong slot; the
strong-entangled route's residual decodability ($0.589$) matches its \emph{raw input's} linear
decodability ($0.58$)---the model adds no disentanglement.}
\label{tab:anatomy}
\end{table}

\begin{figure}[t]
\centering
\includegraphics[width=0.9\textwidth]{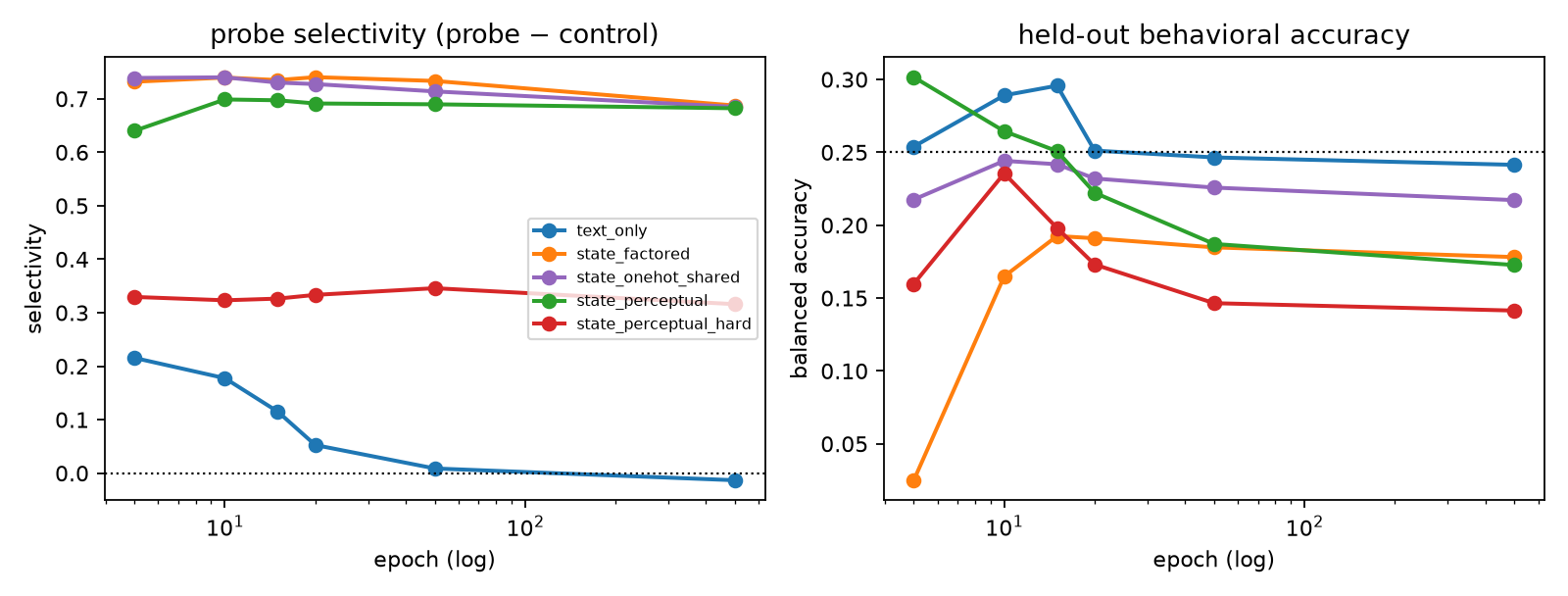}
\caption{Failure anatomy across training ($n{=}10$; log-scale epochs; the last point is the
converged model). Left: control-subtracted probe selectivity for the correct held-out attribute at
the readout position. Right: held-out behavioral accuracy. The symbolic route (blue) briefly
represents the answer exactly while it briefly uses it, then loses both; index routes
(orange/purple) keep the answer decodable throughout while behavior stays at or below chance; the
entangled route (red) is pinned near its input's raw linear readability.}
\label{fig:anatomy}
\end{figure}

Three distinct failure modes emerge (Figure~\ref{fig:anatomy}). \emph{Representation death}
(symbolic): the answer is weakly
present at the readout position exactly while the behavioral transient lasts, then fades;
converged behavior is near-unbiased guessing among shapes. \emph{Systematic mis-binding} (index
routes): the answer remains near-perfectly linearly decodable in the residual stream at every
checkpoint, yet the converged model reads the wrong slot on up to $87\%$ of discriminating states---an intact representation but a behavioral output that depends on the wrong slot, the mechanistic ground of the
below-chance entries in Table~\ref{tab:zeroshot}. \emph{No readability gained} (strong-entangled):
residual decodability ($0.589$) equals the input's raw decodability at every checkpoint; the model
inherits its input's readability rather than improving on it. (We stop short of ``the information
is present but unused'': the strong code is injective---exact ceiling $1.0$, \S\ref{sec:ceilings}---so
no information is lost, but a nonlinear MLP decoder still recovers its factors at only $0.56$ from
the enumerated states, so it is hard for the decoder \emph{and} model classes we test, not merely
for a linear probe.)

\paragraph{From decodable to causal.} Probes show \emph{what is present}, not \emph{what is used}.
We therefore intervene directly on the world state and read the output (Table~\ref{tab:causal}). On
the held-out ``shape of right'' query, varying the correct (right) slot leaves the output almost
unchanged (all routes $0.15$--$0.24$, i.e.\ chance), while varying the wrong (left) slot moves it
strongly (index routes $0.71$--$0.86$; symbolic $0.51$)---the converged output is more sensitive to
the wrong object. As a positive control, on a \emph{trained} query type the same intervention shows
every route showing substantially greater causal dependence on the queried slot ($0.67$--$1.00$)
than on the unqueried one ($\approx 0.25$). What the intervention adds over the probe is real but bounded: it rules out
``decodable-but-causally-inert'' and shows the output functionally depends on the wrong slot under
input intervention, with the answer-representation intact. It identifies the functional input dependence of the converged output: it
reuses the readout for the trained shape slot rather than the held-out one (holding out ``shape of
left'' instead flips which slot is read). The claim is the intact-representation-but-wrong-slot-dependence
dissociation, not the direction of the error.

\begin{table}[h]
\centering\small
\begin{tabular}{lcc|cc}
\toprule
& \multicolumn{2}{c|}{Held-out ``shape of right''} & \multicolumn{2}{c}{Trained ``shape of left''}\\
Route & vary right (correct) & vary left (wrong) & vary left (correct) & vary right\\
\midrule
\texttt{text\_only} & 0.24 & 0.51 & 1.00 & 0.25\\
\texttt{state\_factored} & 0.18 & 0.71 & 1.00 & 0.25\\
\texttt{state\_onehot\_shared} & 0.22 & \textbf{0.86} & 1.00 & 0.25\\
\texttt{state\_perceptual} & 0.17 & 0.43 & 0.67 & 0.26\\
\texttt{state\_perceptual\_hard} & 0.15 & 0.30 & 0.89 & 0.25\\
\bottomrule
\end{tabular}
\caption{Causal input-intervention on the held-out binding type ($n{=}10$; chance $0.25$). Each
number is the fraction of (background, value) cases in which the output equals the shape set in the
varied slot. On the held-out query the output is insensitive to the correct slot and, especially for
the index routes, depends more on the wrong slot; on a trained query it depends much more on the
queried slot than on the unqueried one---a direct manipulation of the input, with no probe
assumptions.}
\label{tab:causal}
\end{table}

\section{Results IV: Readability Governs Few-Shot Binding}
\label{sec:fewshot}
\begin{figure}[t]
\centering
\includegraphics[width=0.62\textwidth]{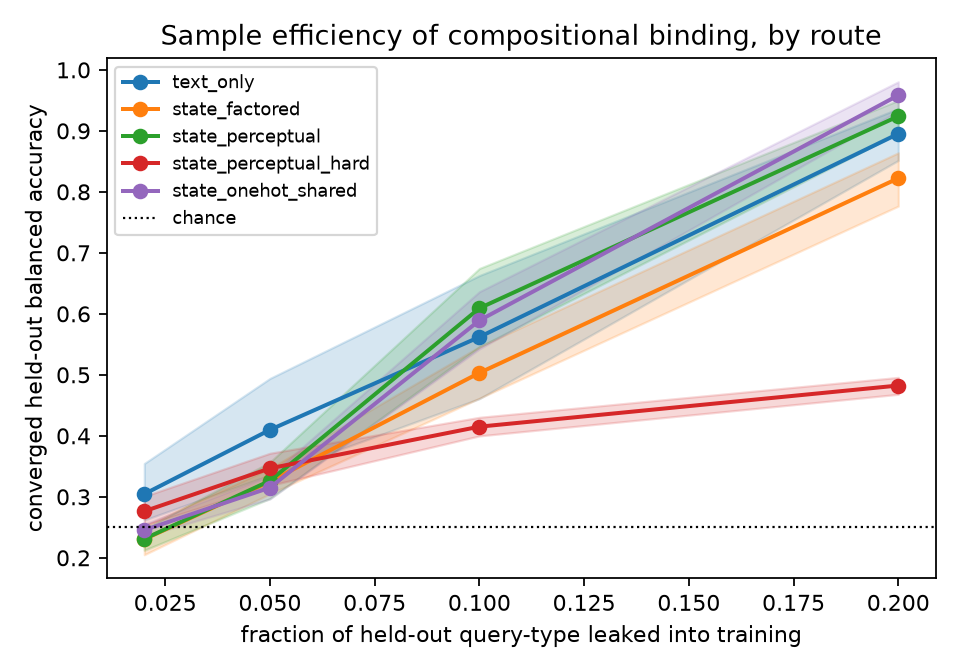}
\caption{Dose-response of compositional binding ($n{=}20$ seeds, converged accuracy, exploratory
holdout; bands are bootstrap 95\% CIs). The entangled route (red) saturates far below
everything, which is the readability effect. The readable routes (blue, green, purple) also sit
above the modular oracle (orange), but that ordering is specific to the broadcast-additive injection
used here and does not survive prefix routing (\S\ref{sec:prefix}); v1 read it as a pathway effect
and we withdraw that reading.}
\label{fig:dose}
\end{figure}

Zero-shot composition fails everywhere, but the cliff is informative once graded: we leak a
fraction $f \in \{0.02, 0.05, 0.1, 0.2\}$ of the held-out query type into training and measure
converged accuracy on the rest (Figure~\ref{fig:dose}). All readable routes rise smoothly from chance to
$0.82$--$0.96$ at $f{=}0.2$, while the strong-entangled route saturates near $0.48$; the question
becomes \emph{which pathway properties govern the rate}.

\paragraph{Entanglement does not explain few-shot transfer.} We initially hypothesized that \emph{entangled} routes
induce compositional structure (the model must compute factors, so it builds factorized machinery
that transfers). The strong-entangled route is this hypothesis's adversarial test: if entanglement
drives composition, harder entanglement should help. It does the opposite---worst pooled over doses,
saturating near $0.48$ while every readable route keeps rising past $0.82$. We score this
against the \emph{one-hot-shared} route rather than the factored oracle: the oracle is the one route
whose level moves with the injection site (\S\ref{sec:prefix}), which makes it an unsafe reference
for anything else. Pooled disadvantage $-0.147$ $[-0.163,-0.132]$, $p = 1.9\times10^{-6}$
(exploratory) and $-0.175$ $[-0.197,-0.153]$, $p = 1.9\times10^{-6}$ (confirmatory); against the
oracle, as v1 reported it, $-0.090$ and $-0.112$. Is the strong code's information merely linearly hidden? No: a nonlinear MLP
decoder recovers its factors at only $0.56$ from the enumerated states (vs.\ $0.95$/$0.98$ linear/MLP
for the weak code), so the $d{=}8$ saturating mix is poorly readable in the tested probe classes, not
just to a linear probe, and the trained model's residual readability ($0.59$) sits near this generic-decoder performance
($0.56$--$0.58$). We therefore state a narrower conclusion: a code decoded only weakly by our linear
and MLP probes is the code the model binds through worst, and the model never reads a code better
than a generic decoder does. We do \emph{not} claim ``linear readability gates binding even when
information is nonlinearly recoverable''---the decisive high-information, linearly-unreadable,
MLP-recoverable cell is not one we could construct with a fixed tanh mix (the compression that
reduced linear readability also reduced recoverability by our MLP decoder), and building it is open
work.

\paragraph{Readability, not input dimension, and graded.} The strong route is lower-dimensional
($d{=}8$) than the readable routes ($d{=}16$), raising the confound that its few-shot deficit
reflects input dimension rather than readability, and the worry that we simply built one
adversarially unreadable code. Two controls answer both. First, a dimension-matched control: a
$d{=}16$ entangled code (higher gain and depth) that is injective (exact ceiling $1.0$) yet poorly
readable (raw linear $0.57$, matching the $d{=}8$ code's $0.58$) pools $-0.294$ ($p=0.002$, $n{=}10$)
below the $d{=}16$ readable code and, if anything, sits slightly below the $d{=}8$ strong code
($-0.105$). Second, a graded readability sweep at fixed $d{=}16$: four shared-projection codes at raw
linear decodability $\{0.95, 0.82, 0.76, 0.53\}$ produce \emph{monotonically} decreasing few-shot
accuracy (Table~\ref{tab:readsweep}; Spearman $\rho{=}1.0$ at both $f{=}0.1$ and $0.2$, $n{=}10$).
Few-shot binding is thus a graded function of readability at fixed dimension, not an artifact of one
code---readability, not input dimension, drives the deficit.

\begin{table}[h]
\centering\small
\begin{tabular}{lcccc}
\toprule
Shared-projection code ($d{=}16$, injective) & \multicolumn{4}{c}{decreasing readability $\rightarrow$}\\
Raw linear decodability & $0.95$ & $0.82$ & $0.76$ & $0.53$\\
\midrule
Few-shot binding accuracy, $f{=}0.1$ & $0.57_{[.49,.65]}$ & $0.48_{[.46,.51]}$ & $0.47_{[.44,.49]}$ & $0.32_{[.28,.35]}$\\
Few-shot binding accuracy, $f{=}0.2$ & $0.93_{[.89,.96]}$ & $0.67_{[.64,.70]}$ & $0.57_{[.52,.62]}$ & $0.34_{[.31,.37]}$\\
\bottomrule
\end{tabular}
\caption{Graded readability sweep at \emph{fixed} input dimension ($d{=}16$, shared projection, all
four codes injective with exact ceiling $1.0$; $n{=}10$, chance $0.25$; subscripts are $95\%$
bootstrap CIs). Few-shot binding decreases monotonically with raw linear decodability (Spearman
$\rho{=}1.0$ at both doses), so the
strong-entangled deficit is a graded function of readability, not an artifact of a single
adversarially-unreadable code or of input dimension.}
\label{tab:readsweep}
\end{table}

\paragraph{Among readable routes the clean oracle loses---but not because of sharing.} The
remaining ordering is that the per-factor oracle is the \emph{least} sample-efficient readable
route: pooled over doses, one-hot-shared beats it by $+0.057$ ($p = 0.0020$) exploratory and
$+0.062$ ($p = 1.1\times10^{-4}$) confirmatory, weak-perceptual by $+0.052$\,/\,$+0.084$, and the
symbolic route by $+0.073$\,/\,$+0.061$ ($p = 0.027$\,/\,$0.053$, directional in confirmation),
while one-hot-shared and weak-perceptual are indistinguishable ($p = 0.52$\,/\,$0.083$;
Table~\ref{tab:fewshot}). That the cleanest disentangled code is not the best route is itself an
instance of disentanglement-is-not-sufficient \citep{montero2021role}---it also yields no zero-shot
composition (\S\ref{sec:endpoint}). v1 went further and attributed the ordering to
\emph{input-pathway parameter sharing}: a shared projection receives gradients from every query type
and so already implements the held-out one, while private per-factor tables do not.
\textbf{We withdraw that attribution.} Three checks, run after v1, retire it in turn.

\paragraph{(i) The compared cells are one function class, so sharing was never manipulated.} The
one-hot state is four disjoint blocks of a $16$-dimensional vector, so a shared
$\mathrm{Linear}(16,24)$ applied to it decomposes exactly into four per-factor matrices acting on
their own blocks. The shared, untied, and per-factor pathways we compared therefore implement the
same set of functions ($408/480/384$ trainable pathway parameters; the shared pathway is the
per-factor one plus a single $24$-dimensional bias), and the oracle-vs.-one-hot-shared contrast
varies no sharing parameter---the weight-tying intuition we imported \citep{csordas2021devil} does
not apply to these cells at all. Manipulating sharing \emph{directly}, by routing the same one-hot
code through a shared projection and through per-factor tables, gives nothing: $-0.024$
$[-0.071,+0.023]$, $p = 0.38$ ($n{=}30$) at width $24$, and $+0.075$ and $+0.007$ at widths $12$ and
$48$, inconsistent in sign.

\paragraph{(ii) What differs is the initialization scale of the injected state.} The routes differ
in fan-in, hence in how large the injected state is relative to the token and position embeddings it
is added to. At initialization, over all $256$ states and $200$ seeds, the ratio
$\lVert\text{state}\rVert/\lVert\text{token}+\text{pos}\rVert$ at the readout position is $1.59$ for
the oracle, $0.40$ strong-entangled, $0.26$ one-hot-shared and $0.20$ weak-perceptual: the oracle
injects a signal several times larger than everything else at that position, and pre-norm
LayerNorm then divides the budget by that ratio. Isolating scale reproduces the effect. A per-factor
pathway with its bias removed is function-class identical to the oracle and differs from it only in
injected scale, and it beats the oracle by $+0.092$ $[+0.043,+0.138]$, $p = 0.0013$ ($n{=}30$), and
by $+0.096$ and $+0.155$ at widths $12$ and $48$, significant at each. Halving the oracle's injected
state ($\sigma{=}0.5$) recovers $+0.043$ $[+0.015,+0.072]$, $p = 0.0084$ ($n{=}70$, three doses).

\paragraph{(iii) And that scale mechanism belongs to the injection site, not to the pathway.} Both
accounts above presuppose that the state is \emph{added} to every token embedding, which is what
makes its norm compete with the token and position signal. Giving the state its own sequence
position instead removes the competition---and removes the effect. In a matched prefix sweep ($20$
seeds/cell paired by initialization seed, same splits, budget and doses) the oracle is the route that
moves, rising from $0.470$ to $0.519$ pooled, while one-hot-shared goes from $0.527$ to $0.512$; the
one-hot-over-oracle advantage falls from $+0.057$ to $-0.006$ $[-0.045,+0.033]$, $p = 0.81$, and the
$\sigma{=}0.5$ recovery from $+0.043$ to $+0.006$ $[-0.020,+0.034]$. \S\ref{sec:prefix} reports the
sweep in full, with the check that this is not a training failure. We therefore make no claim that
input pathways order the readable routes. What the comparison does establish is narrower and, we
think, more useful: \emph{where} conditioning information enters a transformer can by itself produce
an apparent input-pathway effect, so a route comparison has to hold the injection site fixed as well
as the information content.

\begin{table}[h]
\centering\small
\begin{tabular}{lcc|cc|cc}
\toprule
& \multicolumn{2}{c|}{Exploratory (\texttt{bind:3})} & \multicolumn{2}{c|}{Confirmatory
(\texttt{bind:1})} & \multicolumn{2}{c}{Prefix (\texttt{bind:3})}\\
Contrast & pooled $\Delta$ & $p$ & pooled $\Delta$ & $p$ & pooled $\Delta$ & $p$\\
\midrule
\multicolumn{7}{l}{\emph{vs.\ the factored oracle: v1's factor one, specific to additive injection}}\\
\texttt{state\_onehot\_shared} & $+0.057$ & $0.0020$ & $+0.062$ & $1.1\times10^{-4}$ & $-0.006$ & $0.81$\\
\texttt{state\_perceptual} (weak) & $+0.052$ & $0.0042$ & $+0.084$ & $2.7\times10^{-5}$ & $-0.031$ & $0.083$\\
\texttt{text\_only} & $+0.073$ & $0.027$ & $+0.061$ & $0.053$ & --- & ---\\
\texttt{state\_perceptual\_hard} & $-0.090$ & $3.8\times10^{-6}$ & $-0.112$ & $5.7\times10^{-6}$ & $-0.153$ & $1.9\times10^{-6}$\\
one-hot vs.\ weak-perceptual & $+0.004$ & $0.52$ & $-0.022$ & $0.083$ & --- & ---\\
\midrule
\multicolumn{7}{l}{\emph{readability, scored against the one-hot route: the surviving factor}}\\
strong-entangled vs.\ one-hot & $-0.147$ & $1.9\times10^{-6}$ & $-0.175$ & $1.9\times10^{-6}$ & $-0.147$ & $1.3\times10^{-5}$\\
\bottomrule
\end{tabular}
\caption{Few-shot binding, pooled contrasts. \emph{Pooled} $\Delta$ is the per-seed mean of the
dose-wise difference in balanced accuracy across $f\in\{0.02,0.05,0.1,0.2\}$; $p$ is an exact paired
Wilcoxon signed-rank test of that per-seed quantity against $0$ ($n{=}20$ per cell, paired by
initialization seed). The \emph{Prefix} columns repeat the exploratory design with the state placed
at its own sequence position instead of added to every token embedding (\S\ref{sec:prefix});
\texttt{text\_only} has no injected state and hence no prefix arm. \textbf{Top panel:} every
route-vs-oracle advantage that v1 attributed to input-pathway parameter sharing is absent under
prefix routing, and the reason is that the oracle is the route that moves (pooled $0.470 \to
0.519$)---which is also why the oracle is not a safe reference for anything else. Within the
additive family all four comparisons survive Holm--Bonferroni (exploratory adjusted $p\le0.027$;
confirmatory $p\le0.053$, the symbolic route directional). \textbf{Bottom panel:} readability,
scored against the most readable route, is the same in both injection schemes. Pooled Cliff's
$\delta$ for the additive arms (exploratory\,/\,confirmatory): one-hot $0.33/0.38$, weak-perceptual
$0.38/0.55$, symbolic $0.33/0.20$, strong-entangled $-0.15/-0.47$. The strong-entangled deficit
grows with dose as the readable routes climb and it saturates.}
\label{tab:fewshot}
\end{table}

\paragraph{Larger world.} To check the account is not a two-object artifact we rerun the
k-shot sweep in a three-object world ($729$ states, six query types, chance $0.333$; $n{=}10$). The
readability effect carries over: the strong-entangled route again collapses (pooled $-0.348$ against
the oracle, $p = 0.002$, saturating near $0.51$ while readable routes exceed $0.94$). Two effects do
not survive as pathway claims. The weak-perceptual route's small edge over the oracle does not
replicate at all (pooled $-0.001$, $p = 0.77$), and the shared one-hot pathway's edge over the
oracle, which does replicate here (pooled $+0.066$, $p = 0.037$), is the additive-injection artifact
of \S\ref{sec:prefix} rather than a property of the pathway; in hindsight the weak-perceptual
non-replication was the first sign that the ordering among readable routes was fragile. Readability
is the portable factor; the ranking among readable codes is not.

\paragraph{Two controls against artifacts.} \emph{Description order.} The symbolic route's edge is
not an absolute-position shortcut: shuffling the two objects while tagging each with its position
word (so binding must use the tag, not token position) leaves few-shot efficiency unchanged
(free-order $-$ fixed-order $= +0.005$, $p = 1.0$; $n{=}10$) and still above the oracle ($+0.109$,
$p = 0.049$). \emph{Mix instantiation.} Because the ``perceptual'' routes rest on one fixed
synthetic map, we re-instantiate each with three independent random mixes: at $f{=}0.1$ all three
weak mixes beat the oracle (mean $0.553$ vs.\ $0.480$) and all three strong mixes fall below it
(mean $0.382$), so the readability effect is a property of the code class, not of one lucky matrix.

\paragraph{The double dissociation, withdrawn.} v1 paired the trajectory result with the
few-shot one: one-hot-shared has the \emph{highest} zero-shot mis-binding ($0.866$) and essentially
no transient (collapse $0.034$) yet is among the most few-shot-efficient, while the strong-entangled
route has a large transient ($0.140$) yet the worst few-shot efficiency. Both legs have since failed:
the transient leg rests on a selection statistic (\S\ref{sec:dynamics}), and the few-shot leg
attributed the ordering among readable routes to pathway sharing. What survives is weaker but still
worth stating: zero-shot failure mode does not predict few-shot capability---the route with the
worst mis-binding is not the worst few-shot learner, and the worst few-shot learner is identified by
readability instead.

\section{Discussion}
\label{sec:discussion}
\paragraph{What this says about grounding claims.} Delivering a provably sufficient world state
through a non-symbolic channel bought, at this scale: no zero-shot composition, a few-shot
advantage only when the code is nearly linearly readable, and a penalty when entanglement makes the
code poorly readable. The operative variable was mundane---the practical readability of a fixed \emph{synthetic}
perceptual code (not a learned or naturalistic one)---not ``groundedness.'' And the one apparent
pathway effect we did find turned out to belong to the injection site rather than the pathway. We
suggest route-based grounding claims in larger systems be held to the control this implies: match
information exactly, hold fixed \emph{where} the information enters, and only then show that the
route effect survives.

\paragraph{What it says about tiny-model interpretability.} The regime is a feature: exhaustive
evaluation turned a trajectory into a measurable object, exact ceilings turned ``failure'' into a
classification (information- vs.\ inductive-bias-limited), and cheap exhaustive runs made it
possible to re-test our own mechanism claim two ways---against a selection-free statistic and
against a second injection site---on the same manifests. That is also the cautionary half of the
story: a parameter-matched extra condition looked like it settled a two-way mechanism question, and
it did not, because the two cells turned out to lie in one function class and to differ in a
variable we had not thought to control. The cost is scale; none of our statements should be
extrapolated beyond small transformers without new evidence.

\paragraph{Limitations.} Worlds of $128$--$729$ states and $1$--$2$-layer models. Entanglement is
operationalized by two main routes plus a dimension-matched low-readability control; the strong code is \emph{injective} (exact ceiling $1.0$, so no
information is destroyed) but hard for both linear and MLP decoders ($0.58$/$0.56$), so we cannot
cleanly isolate ``linearly hard'' from ``hard for every decoder and model we tested''---the safe
reading is that practical readability, not information content, gates binding here; whether a
high-information code that is linearly unreadable but nonlinearly recoverable would bind well is an
untested cell. Our causal test intervenes on the input (a clean but coarse handle); internal
readout-swap or activation-patching evidence would localize this dependence to an internal readout. The symbolic
route's fixed description order permits absolute-position shortcuts (its binding may be easier than
free-order text). Few-shot leakage varies which examples leak by seed but not the schedule. The
readability effect replicated on a second held-out type, in a three-object world and under both
injection schemes, but both held-out types query shape, so attribute- and modality-general
replication is future work. The ordering among readable routes does not replicate across injection
sites and we draw no conclusion from it; since the only two sites we tried are broadcast addition
and a dedicated prefix position, we also cannot say which conditioning site a larger model's
behaviour would resemble. Finally, the
confirmatory arm fixed predictions and configurations before running, but exploration and
confirmation were performed by the same authors on the same testbed.

\paragraph{Future work.} Freeing description order to force genuine symbolic binding; scaling the
world (more objects/factors) to put compositional pressure on the lookup solution; a learned
perception module (trading the fixed-mix control for adaptivity); and testing whether readability
predicts adapter- and prompt-style transfer in large models. The injection-site result suggests a
sharper version of that last question: prompt-style conditioning occupies its own positions while
adapter- and bias-style conditioning adds to a residual stream, so the distinction we stumbled into
here has a direct analogue at scale, and it should be varied deliberately rather than inherited from
an implementation.

\subsubsection*{Broader Impact Statement}
This work is foundational analysis on synthetic worlds; we foresee no direct societal impact
beyond clarifying methodology for grounding and interpretability claims.

\subsubsection*{Reproducibility Statement}
Evaluation is exhaustive (no test-set sampling); all runs are logged to JSONL manifests recording
configs, seeds, and full trajectories; every table and figure re-aggregates from those manifests
without retraining, via the scripts we provide. The manifests for the prefix sweep of
\S\ref{sec:prefix} and for the direct manipulations of \S\ref{sec:fewshot} are included on the same
terms, so every correction this version makes can be recomputed from the logs that v1's claims were
drawn from. The code, manifests, per-seed logs, and exact
commands are available at \url{https://github.com/otanl/microground}.
Table~\ref{tab:spec} is the one-page experimental specification (routes, dimensions,
trainable parameters, raw decodability, exact Bayes ceiling, and train/in-distribution/held-out
accuracy). Determinism caveat: runs are CPU-deterministic given the recorded seeds up to library
versioning, which we record.

\bibliographystyle{tmlr}
\bibliography{refs}

\appendix
\section{Exact Ceiling Values}
\label{app:ceilings}
For each (task, split): informative routes (\texttt{text\_only}, \texttt{state\_factored},
\texttt{state\_onehot\_shared}, \texttt{state\_perceptual}, \texttt{state\_perceptual\_hard},
\texttt{scrambled\_state}) have Bayes ceiling $1.000$ everywhere. Floors
(\texttt{text\_minimal}, \texttt{uninformative\_state}): binding holdout $0.250$; binding random
$0.322$; counterfactual random $0.423$; counterfactual transition holdout $0.833$ (degenerate
held-out target distributions). Trained held-out transition accuracy ($0.000$--$0.058$) therefore
falls below the blind ceiling for every route. The ceiling audit includes all seven routes
(the late-added \texttt{state\_onehot\_shared} and \texttt{state\_perceptual\_hard} were verified
separately: ceiling $1.000$ on every split).

\section{Experimental Specification}
\label{app:spec}
Table~\ref{tab:spec} summarizes the binding task (held-out type ``shape of right,''
\texttt{bind:3}) at $n{=}20$. Total parameters differ by $<8\%$ across routes; crucially the input
pathways of \texttt{state\_onehot\_shared} and \texttt{state\_perceptual} are the identical
$16{\to}24$ linear map (the shared-vs-modular contrast is not a parameter-count artifact). Every
informative route's exact Bayes ceiling is $1.0$; every route memorizes the trained query types
(train column) yet fails the held-out one.

\begin{table}[h]
\centering\small
\resizebox{\textwidth}{!}{%
\begin{tabular}{lccccccc}
\toprule
Route & Params & State dim & Raw lin.\ dec. & Bayes ceiling & Trained-type & All-space & Held-out\\
\midrule
\texttt{text\_only} & 5736 & (tokens) & --- & 1.000 & ${\approx}1.0$ & 0.81 & 0.227\\
\texttt{state\_factored} & 6120 & 4 idx & 1.00 & 1.000 & ${\approx}1.0$ & 0.80 & 0.198\\
\texttt{state\_onehot\_shared} & 6144 & 16 & 1.00 & 1.000 & ${\approx}1.0$ & 0.80 & 0.215\\
\texttt{state\_perceptual} & 6144 & 16 & 0.95 & 1.000 & ${\approx}1.0$ & 0.81 & 0.224\\
\texttt{state\_perceptual\_hard} & 5952 & 8 & 0.58 & 1.000 & $0.885$ & 0.70 & 0.146\\
\bottomrule
\end{tabular}}
\caption{One-page experimental specification (binding holdout, $n{=}20$). ``Trained-type'' is
accuracy on the three query types seen in training (memorized, ${\approx}1.0$; strong-entangled
${\approx}0.9$, still fitting); ``All-space'' is balanced accuracy over all four types
(${=}\,\tfrac14(3{\cdot}\text{trained}+\text{held-out})$); ``Held-out'' is converged balanced
accuracy on the never-trained query type (chance $0.25$).
Per-seed values, all splits, and the full route/split/probe details are in the released manifests
and code.}
\label{tab:spec}
\end{table}

\section{Larger-World Stress Check (Three Objects)}
\label{app:bind3}
Table~\ref{tab:bind3} gives the three-object k-shot sweep ($729$ states, six query types, chance
$0.333$; $n{=}10$) referenced in \S\ref{sec:fewshot}. The shared one-hot pathway leads and the
strong-entangled route collapses at both doses; the weak-perceptual edge over the oracle does not
carry over (it is testbed-specific).

\begin{table}[h]
\centering\small
\resizebox{\textwidth}{!}{%
\begin{tabular}{lccccc}
\toprule
Dose $f$ & \texttt{text\_only} & \texttt{state\_factored} & \texttt{onehot\_shared} & \texttt{perceptual} & \texttt{perceptual\_hard}\\
\midrule
$0.05$ & $0.68_{[.53,.83]}$ & $0.72_{[.66,.79]}$ & $\mathbf{0.81}_{[.74,.87]}$ & $0.71_{[.67,.76]}$ & $0.45_{[.42,.48]}$\\
$0.10$ & $0.94_{[.87,.98]}$ & $0.93_{[.90,.96]}$ & $\mathbf{0.98}_{[.97,.99]}$ & $0.94_{[.93,.96]}$ & $0.51_{[.49,.54]}$\\
\bottomrule
\end{tabular}}
\caption{Three-object world, converged held-out balanced accuracy by route and dose ($n{=}10$,
chance $0.333$; subscripts are $95\%$ bootstrap CIs). Pooled vs.\ oracle: one-hot $+0.066$
($p{=}0.037$); strong-entangled $-0.348$ ($p{=}0.002$); weak-perceptual $-0.001$ ($p{=}0.77$,
non-replicating).}
\label{tab:bind3}
\end{table}

\section{Injection Site: Prefix Routing}
\label{sec:prefix}
Every result in this paper other than the ones in this section conditions the model by
\emph{broadcast addition}: the state embedding is added to the embedding of every token,
$x \leftarrow x + \mathrm{se}$. We inherited that from the testbed's first implementation and never
examined it. Under pre-norm LayerNorm it has a consequence. The injected state and the
token-plus-position signal share a single normalized budget at each position, so a route whose state
embedding is larger at initialization takes more of that budget; and the routes differ in fan-in and
therefore in exactly that norm (\S\ref{sec:fewshot}). Injected scale is thus confounded with
pathway identity throughout the additive arm.

\paragraph{The sweep.} We re-ran the dose-response design with the state given its own sequence
position instead, $x \leftarrow [\,\mathrm{se}\,;x\,]$, holding everything else fixed: the same four
leak fractions $f \in \{0.02, 0.05, 0.1, 0.2\}$, the same exploratory held-out query type, the same
optimizer, budget and split seeds, and $20$ seeds per cell paired with the additive arm by
initialization seed. \texttt{text\_only} injects no state and therefore has no prefix arm.

\begin{table}[h]
\centering\small
\begin{tabular}{lccccc|ccccc}
\toprule
& \multicolumn{5}{c|}{Additive (\S\ref{sec:fewshot})} & \multicolumn{5}{c}{Prefix}\\
Route & $f{=}.02$ & $.05$ & $.1$ & $.2$ & pooled & $f{=}.02$ & $.05$ & $.1$ & $.2$ & pooled\\
\midrule
\texttt{state\_factored} & 0.230 & 0.325 & 0.503 & 0.822 & \textbf{0.470} & 0.264 & 0.331 & 0.581 & 0.899 & \textbf{0.519}\\
\texttt{state\_onehot\_shared} & 0.245 & 0.315 & 0.589 & 0.958 & 0.527 & 0.245 & 0.346 & 0.593 & 0.865 & 0.512\\
\texttt{state\_perceptual} & 0.231 & 0.326 & 0.609 & 0.924 & 0.523 & 0.248 & 0.324 & 0.515 & 0.862 & 0.487\\
\texttt{state\_perceptual\_hard} & 0.276 & 0.347 & 0.415 & 0.482 & 0.380 & 0.258 & 0.327 & 0.412 & 0.466 & 0.366\\
\bottomrule
\end{tabular}
\caption{Converged few-shot binding accuracy by route, dose and injection site ($n{=}20$ seeds per
cell, chance $0.25$, converged $=$ mean over the final $10\%$ of training). Bold marks the factored
oracle's pooled level, the one entry that moves materially between the two schemes. Contrasts and
tests appear in Table~\ref{tab:fewshot}.}
\label{tab:prefix}
\end{table}

\paragraph{Reading.} Three things are visible. \emph{(1)~The oracle is the route that moves.} Its
converged accuracy rises from $0.470$ to $0.519$ pooled over doses ($0.503 \to 0.581$ at $f{=}0.1$),
while one-hot-shared barely changes ($0.527 \to 0.512$ pooled, $0.589 \to 0.593$ at $f{=}0.1$).
Giving the state its own position removes a handicap that was specific to the route whose injected
norm was largest. \emph{(2)~Every route-over-oracle advantage therefore disappears}, including the
one-hot-shared advantage that v1 read as input-pathway parameter sharing ($+0.057 \to -0.006$,
$[-0.045,+0.033]$, $p = 0.81$) and the $\sigma{=}0.5$ scale recovery ($+0.043$, $[+0.015,+0.072]$,
$p = 0.0084$, $n{=}70$ $\to$ $+0.006$, $[-0.020,+0.034]$). \emph{(3)~Readability is untouched:}
scored against the one-hot route, the strong-entangled penalty is $-0.147$ in both schemes
($[-0.163,-0.132]$ additive, $[-0.191,-0.103]$ prefix). The zero-shot endpoint is likewise unchanged
(\S\ref{sec:endpoint}).

\paragraph{Not a training failure.} The prefix arm fits the trained query types as well as the
additive arm: all-space accuracy at $f{=}0.1$ is within $0.02$ of the additive value for every route
($0.888 \to 0.906$ oracle, $0.908 \to 0.908$ one-hot-shared, $0.912 \to 0.891$ weak-perceptual,
$0.780 \to 0.779$ strong-entangled). The advantage that disappears is not lost to a failure to learn
under the new routing.

\paragraph{What we take from this.} We do not claim prefix routing is the right choice. It is a
second arbitrary choice, and the defensible statement is the conditional one: the ordering among
readable routes depends on the injection site, while readability and the zero-shot endpoint do not.
That is enough to retire the pathway claim, and it leaves a methodological point we would not
otherwise have had. In a comparison of input pathways, the site at which conditioning information
enters is a free variable that can produce an apparent pathway effect by itself, so it belongs in
the design and in the report, not in the defaults.

\section{Training and Probing Details}
\label{app:training}
\paragraph{Model.} A pre-norm decoder-only transformer: token embedding plus a fixed sinusoidal
positional embedding, $L$ blocks (multi-head self-attention with a causal mask, $4$ heads; a
two-layer GELU MLP of width $2\times$ hidden), a final LayerNorm, and a linear token head over the
vocabulary read at the last position. Default hidden size $24$, $L{=}1$ ($L{=}2$ in the capacity
arm); ${\sim}6$K parameters (Table~\ref{tab:spec}). When a state channel is present, its embedding
(a sum of per-factor index embeddings, or a linear projection of the real-valued code) is added to
\emph{every} token position.
\paragraph{Optimization.} AdamW ($\beta=(0.9, 0.999)$; weight decay $0.01$ except in the
weight-decay arm), learning rate $10^{-3}$ (except the LR arm), full-shuffle minibatches of size
$32$, cross-entropy on the final-token logits against the single answer token. Epochs: $500$
(binding), $1000$ (counterfactual), $4000$ (weight-decay arm); the exhaustive query set is evaluated
every $5$ epochs, and the primary metric averages the last $10\%$ of evaluations. Seeds set torch,
numpy, and the shuffle generator; initialization and split seeds are separated.
\paragraph{Splits.} Random splits partition the $128$/$256$ states $80/20$ by a split seed. The
binding holdout removes one of the four (position, attribute) query types entirely; its $k$-shot
variant leaks a uniformly random fraction $f$ of that type's states back into training (seeded). The
transition holdout removes a uniformly random $30\%$ of (factor, source-value) successor transitions.
\paragraph{Probes.} Linear probes are 5-fold stratified-CV logistic regression
(\texttt{max\_iter}$=2000$) on the final-token residual; the control task \citep{hewitt2019designing}
fits the same probe to fixed random labels with matched class counts, and we report selectivity
$=$ probe $-$ control. The nonlinear decoder is an MLP (\texttt{hidden}$=(64,64)$,
\texttt{max\_iter}$=2000$) under the same protocol; widening it to $(256,256)$ or deepening to
$(128,128,128)$ did not improve recovery of the strong code ($0.56 \to 0.55 \to 0.54$). ``Raw
decodability'' applies the same probes to the raw state input rather than the residual.

\section{Additional Robustness Detail}
\label{app:robust}
Capacity grid (hidden, layers; parameters): $(24,1;\,6.1\text{K}), (48,1;\,20.7\text{K}),
(48,2;\,39.6\text{K}), (96,2;\,153\text{K})$; $n{=}8$--$20$ seeds each;
converged held-out binding by route remains within $[0.15, 0.24]$ throughout (the strong-entangled
route reaches $0.146$). Learning-rate arm:
$\{3\times10^{-4}, 3\times10^{-3}\}$ on the binding holdout (symbolic, oracle, weak-perceptual;
$n{=}10$): converged accuracy $0.18$--$0.24$, no significant route differences. Weight-decay grid on
transitions: $\{0.01, 0.1, 0.3, 1.0\} \times 4{,}000$ epochs, $n{=}5$ each; held-out accuracy
$0.000$ in all $20$ cells while train-side accuracy is flat at memorization from
$\sim$epoch~40. \emph{Dimension-matched readability control} (\S\ref{sec:fewshot}): a $d{=}16$
route from a fixed 3-layer tanh mix with higher gain through the same $16{\to}24$ shared pathway as
the weak-perceptual route; exact Bayes ceiling $1.0$, raw linear decodability $0.57$ (matching the
$d{=}8$ code's $0.58$), an injective code verified over all $256$ states. On the binding holdout
($n{=}10$), its pooled few-shot deficit versus the $d{=}16$ readable weak-perceptual route is
$-0.294$ ($p=0.002$), and it sits at or below the $d{=}8$ strong code ($-0.105$), so the deficit
tracks readability rather than input dimension. \emph{Graded readability sweep} (\S\ref{sec:fewshot}):
four shared-projection $d{=}16$ codes (all injective, exact ceiling $1.0$) at raw linear decodability
$\{0.95, 0.82, 0.76, 0.53\}$ give few-shot accuracy $\{0.57, 0.48, 0.47, 0.32\}$ at $f{=}0.1$ and
$\{0.93, 0.67, 0.57, 0.34\}$ at $f{=}0.2$ ($n{=}10$; Spearman $\rho{=}1.0$ at both doses), so few-shot
binding is monotone in readability at fixed dimension. A capacity sweep of the MLP decoder
($(64,64)\to(256,256)\to(128,128,128)$) does not improve recovery of the strong code
($0.56\to0.55\to0.54$). Full per-seed values are in the released manifests.

\end{document}